\pdfoutput=1

\documentclass[11pt]{article}

\usepackage[final]{EMNLP2023}

\usepackage{times}
\usepackage{amsmath}
\usepackage{latexsym}
\usepackage[ocr-a]{ocr}
\usepackage{comment}
\usepackage[T1]{fontenc}

\usepackage[utf8]{inputenc}

\usepackage{microtype}
\usepackage{xspace}
\usepackage{graphicx}
\usepackage{tabularx}
\usepackage{booktabs}
\usepackage{enumitem}

\usepackage{inconsolata}
\newcommand\specialfont[1]{\smash{{\usefont{T1}{}{m}{n}#1}}}
\newcommand{\promptmodel}{\specialfont{Prompt2Model}\xspace}

\title{\textsc{Prompt2Model}: \\ Generating Deployable Models from Natural Language Instructions}

\author{
\bf{Vijay Viswanathan}$^{1}$\thanks{~~equal contribution.}\hspace{0.5em}, \bf{Chenyang Zhao}$^{1,2*}$, \\ \bf{Amanda Bertsch}$^{1}$, \bf{~Tongshuang Wu}$^{1}$, \bf{Graham Neubig}$^{1}$ \\ 
\textsuperscript{1}Carnegie Mellon University, \textsuperscript{2}Tsinghua University}

\begin{document}
\maketitle
\begin{abstract}

Large language models (LLMs) enable system builders today to create competent NLP systems through prompting, where they only need to describe the task in natural language and provide a few examples.
However, in other ways, LLMs are a step backward from traditional special-purpose NLP models; they require extensive computational resources for deployment and can be gated behind APIs.
In this paper, we propose \promptmodel, a general-purpose method that takes a natural language task description like the prompts provided to LLMs, and uses it to train a special-purpose model that is conducive to deployment.
This is done through a multi-step process of retrieval of existing datasets and pretrained models, dataset generation using LLMs, and supervised fine-tuning on these retrieved and generated datasets.
Over three tasks, we demonstrate that given the same few-shot prompt as input, \promptmodel trains models that outperform the results of a strong LLM, \texttt{gpt-3.5-turbo}, by an average of 20\% while being up to 700 times smaller.
We also show that this data can be used to obtain reliable \emph{performance estimates} of model performance, enabling model developers to assess model reliability before deployment. \promptmodel is available open-source at \url{https://github.com/neulab/prompt2model}.\footnote{Our demo video is posted at \url{youtu.be/LYYQ_EhGd-Q}.}
\end{abstract}

\section{Introduction}

Traditionally, building an NLP model from scratch has been a substantial undertaking.
An NLP practitioner seeking to solve a new problem would need to define their task scope, find or create data that specifies the intended system behavior, choose a suitable model architecture, train the model, assess its performance through evaluation, and then deploy it for real-world usage~\cite{paleyes2022challenges}.

LLMs like GPT-3 \citep{NEURIPS2020_1457c0d6, 10.1145/3560815} offer a lighter-weight paradigm for NLP system construction through ``prompting''~\citep{10.1145/3411763.3451760}. 
Practitioners can now write a prompt specifying the intended system behavior (optionally with a few demonstrations), and ask an LLM to generate a desired output via text completion. 
This makes it possible to prototype NLP systems rapidly for a variety of applications without writing a single line of code \citep{floridi2020gpt}.

\begin{figure}[t]
\centering
\includegraphics[trim={0 18cm 45cm 0cm}, clip, width=1\linewidth]{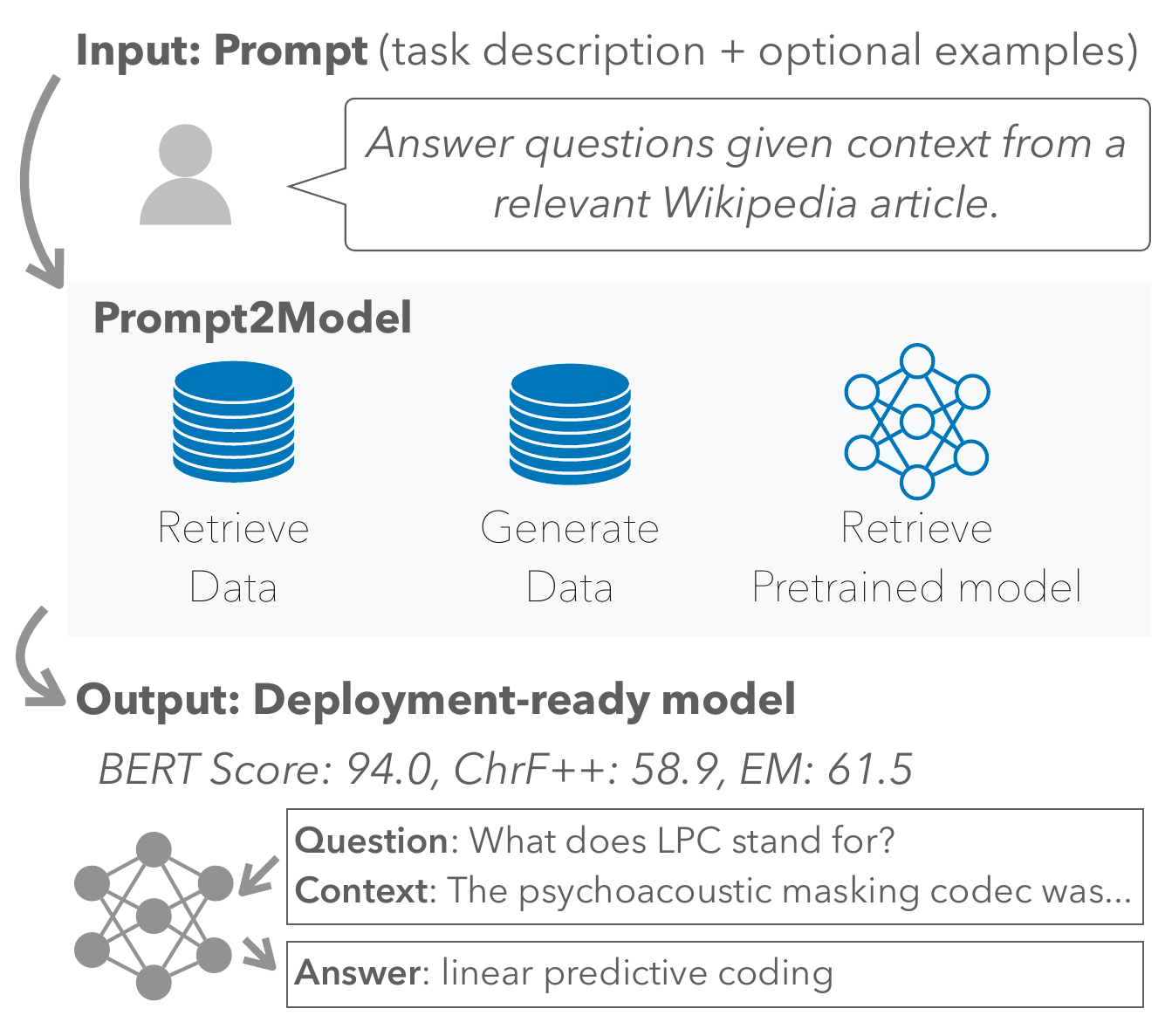}
\vspace{-15pt}
\caption{\promptmodel is a framework for generating a small yet accurate model from a prompt.}
\vspace{-15pt}
\label{fig:intro_diagram}
\end{figure}

However, there is still a gap between proof-of-concept prototyping --- showing LLMs can be prompted for a particular task --- and practical deployment.
Prompting LLMs can be expensive as they require either a significant amount of computing or access to commercial APIs, and their reliance on the input prompt quality makes them unstable compared to trained models \citep{min-etal-2022-rethinking,bubeck2023sparks}.
Because practitioners usually do not have enough annotated validation data to measure their system performance, it is also more challenging for them to debug their systems before deployment~\cite{jiang2022promptmaker}.
Additionally, LLM-prompted systems pose usability challenges. Practitioners have expressed concerns about the high serving cost and slow prediction time associated with using LLMs~\citep{Park2022LUTGEMMQM}, and those working in high-stakes domains cannot rely on commercial LLM APIs due to privacy concerns. For instance, sharing user data with LLM service providers is illegal for many applications in the US~\citep{gpt_hipaa}.

In this work, we present \textbf{\promptmodel}, a system that retains the ability to specify system behavior in a light-weight way through \emph{prompting}, while still resulting in a \emph{deployable special-purpose model}, maintaining all the advantages thereof.
\promptmodel is designed as an automated pipeline that extracts essential task information from users' prompts and then automatically collects and synthesizes task-specific knowledge through three channels:
\begin{itemize}[nosep, leftmargin=1.1em,labelwidth=*,align=left]
\item \emph{Dataset retrieval}: Whenever possible, we collect training data by retrieving task-relevant annotated data~\citep{farber2021recommending, viswanathan-etal-2023-datafinder}.
\item \emph{Dataset generation}: We distill knowledge from an LLM (``teacher model'') by employing it to generate a pseudo-labeled dataset. Prior work has demonstrated that such a dataset can be used to train a smaller ``student'' model to emulate the behavior of the teacher model~\cite{wang-etal-2021-want-reduce, He2023AnnoLLMML, Gudibande2023TheFP}.
\item \emph{Model retrieval}: Based on the prompt, we identify a pretrained language model whose parametric knowledge is appropriate for the user's intent. This chosen model serves as the student model and is further fine-tuned and evaluated using the generated and retrieved data.
\end{itemize}

\promptmodel is designed to support different instantiations of each of these components.
We provide a reference implementation where we demonstrate its utility with a \texttt{gpt-3.5-turbo}-based dataset generator, a dataset retriever based on DataFinder~\cite{viswanathan-etal-2023-datafinder}, and a model retriever using BM25.
We evaluate on three tasks covering both traditional NLP benchmarks and novel applications and find that, empirically, \promptmodel sometimes produces small models that outperform \texttt{gpt-3.5-turbo} when using the same prompt as input. On 2 of these 3 tasks, we observe >20 point improvements over the \texttt{gpt-3.5-turbo} baseline, despite the final model produced by \promptmodel being up to 700 times smaller.
We also find that we can generate effective evaluation datasets; performance improvements on these synthetic clones of real benchmarks also hold on their real counterparts.
We believe that \promptmodel can serve the following purposes for the community:

\begin{enumerate}[nosep, leftmargin=1.1em,labelwidth=*,align=left]
\item \textbf{A tool for quickly building small and competent NLP systems}:  \promptmodel can be directly used to produce task-specific models that outperform LLMs in a few hours without any manual data annotation or architecture design.
The method bridges the gap between the proof-of-concept LLM prototyping and the practical deployment of the model.

\item \textbf{A testbed for end-to-end, prompt-based model training}: 
Given \promptmodel's extensible design, it can offer a platform for exploring new techniques in model distillation, dataset generation, synthetic evaluation, dataset retrieval, and model retrieval. Our platform allows studying these components using extrinsic downstream metrics, enabling empirical progress on these research areas.
\end{enumerate}

\section{Prompt2Model Framework}
\label{sec:prompt2model_framework}
\begin{figure*}[h]
\begin{center}

    \includegraphics[width=0.8\textwidth]{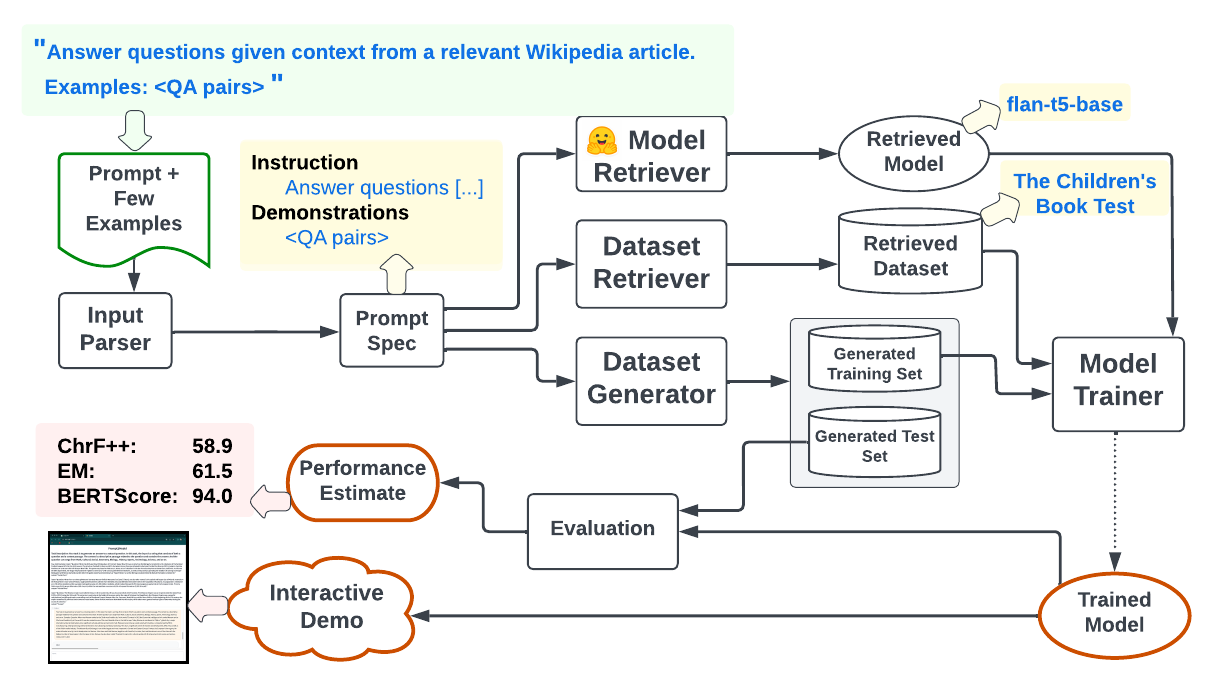}
    
\end{center}
\vspace{-10pt}
\caption{The \promptmodel architecture seeks to automate the core machine learning development pipeline, allowing us to train a small yet accurate model from just a prompt.
}
\label{fig:architecture_diagram}
\vspace{-10pt}
\end{figure*}

Our system, \promptmodel, provides a platform to automate the components of a 
 machine learning pipeline: data collection, model training, evaluation, and deployment. We illustrate our automated pipeline in \autoref{fig:architecture_diagram}. 
At the core is our automatic data collection system, which leverages dataset retrieval and LLM-based dataset generation to obtain labeled data relevant to the user's needs. We then retrieve pretrained models which we finetune on the training splits of the collected datasets. Finally, we evaluate our trained models on the test splits of the same datasets and optionally create a web UI that can be used to interact with the model.

Our general-purpose method is designed to be modular and extensible; each component can be implemented differently or disabled by a practitioner. We give an overview of our framework, then in \autoref{sec:reference_implementation} we describe our reference implementation.

\paragraph{Prompt Parser}
As the primary input to our system, users provide prompts similar to those used for LLMs. These prompts comprise an instruction and, optionally, a few demonstrations of the anticipated behavior. While this open-ended interface is convenient for users, end-to-end ML pipelines may benefit from a \textit{Prompt Parser} that processes this input, such as segmenting the prompt into an instruction and individual demonstrations or translating instructions into English.

\paragraph{Dataset Retriever}
Given a prompt, we first try to discover existing manually-annotated data that can support a user's task description. There are several design decisions for the \textit{Dataset Retriever}:
\begin{enumerate}[nosep, leftmargin=1.2em,labelwidth=*,align=left]
    \item What datasets to search against?
    \item How to index datasets for search?
    \item Which dataset columns are needed for the user's task, and which columns should be ignored?
\end{enumerate}

Prior works by \citet{farber2021recommending} and \citet{viswanathan-etal-2023-datafinder} introduced systems for dataset search. We use the latter, called \emph{DataFinder}, in our implementation, as described in \S\ref{sec:dataset_retriever_impl}.

\paragraph{Dataset Generator}
Not all conceivable tasks have any existing annotated data, and many tasks are only somewhat relevant to an existing dataset. To support a wide range of tasks, we introduce a \textit{Dataset Generator} to produce synthetic training data as per the user-specific requirements parsed by the \emph{Prompt Parser}.
This component presents challenges related to cost efficiency, generation speed, example diversity, and quality control. We discuss our suggested solution to these challenges in \S\ref{sec:dataset_generation_impl}.

\paragraph{Model Retriever}
Besides training data, we must identify an appropriate model to finetune. We cast this as a retrieval problem, where each model is represented by user-generated descriptions and metadata such as popularity or tasks supported. The reference implementation of our \textit{Model Retriever}, described in \S\ref{sec:model_retriever_impl}, searches against pretrained models on Hugging Face \citep{wolf-etal-2020-transformers}, but this could instead cover other model repositories such as Model Zoo \citep{model_zoo}.

\paragraph{Training}
Given retrieved and generated datasets and a pretrained model, we use a \textit{Model Trainer} to finetune the model on a subset of the data. We currently train models by treating all tasks as text-to-text generation \citep{10.5555/3455716.3455856}, as described in \S\ref{sec:training_impl}, but emphasize that this component can be extended in the future to support new approaches.

\paragraph{Evaluation}
After training models on a portion of the retrieved and generated datasets, we give the remaining data to an \textit{Model Evaluator} module. We aim to support a variety of tasks, and selecting the correct task-specific metrics for an arbitrary task is a difficult problem. We describe our suggested strategies for task-agnostic evaluation in \S\ref{sec:eval_impl}.

\paragraph{Web App Creation}
To enable developers to expose a model to collaborators or users, we include an optional component called the \textit{Demo Creator} to create a graphical interface to interact with the model. We briefly describe our implementation of this component in \S\ref{sec:web_app_creator_impl}.

\section{Reference Implementation}
\label{sec:reference_implementation}

\promptmodel is designed modularly to support customization of each component in our framework (described in \S\ref{sec:prompt2model_framework}), but we have provided a reference implementation to enable immediate adoption.

\subsection{Prompt Parser}
We parse the prompt into \texttt{instruction} and \texttt{demonstrations} fields (shown in \autoref{fig:architecture_diagram}), where the instruction represents the primary task or objective and the demonstrations exemplify the desired behavior. To achieve this, we utilize an LLM with in-context learning to segment user prompts, employing the OpenAI \texttt{gpt-3.5-turbo-0613} in our experiments. If the instruction provided is identified to be in a language other than English, we translate it to English using the DeepL API.%
\footnote{\url{https://www.deepl.com/en/docs-api}}

\subsection{Dataset Retriever}
\label{sec:dataset_retriever_impl}
To retrieve datasets for a prompt, we adapt the \emph{DataFinder} system introduced by \citet{viswanathan-etal-2023-datafinder}. By extracting user-generated dataset descriptions for each dataset in Hugging Face Datasets \citep{lhoest-etal-2021-datasets}, we utilize DataFinder's trained bi-encoder retriever to rank the most relevant datasets. Once a relevant dataset is identified, the next step is to determine which columns of the dataset correspond to the input and the desired output specified by the user. As automatically inducing the correct schema for any dataset can be challenging, we adopt a human-in-the-loop approach. We present the top-$k$ datasets, where $k=25$ by default, to the user and allow them to either select the most relevant dataset or to state that none are a good fit for their task. We then ask the user to identify the appropriate columns for input and output from the dataset's schema.

\subsection{Dataset Generator}
\label{sec:dataset_generation_impl}
We carefully engineered our dataset generator to enable speed-optimized generation at a low-cost while creating diverse and high-quality examples. Our strategy comprises the following components:

\paragraph{High-Diversity Few-Shot Prompting} We use automated prompt engineering to generate a diverse dataset. We augment the user-provided demonstration examples with a random sample of previously generated examples to promote diversity and avoid generating duplicate examples. Without this strategy, 120 out of 200 generated QA examples were duplicates; with it, only 25 were duplicates.

\paragraph{Temperature Annealing} We adjust the sampling temperature from low (favoring deterministic outputs) to high (encouraging diverse exploration) proportionally to the number of examples already generated. This modulation helps preserve output quality while gradually encouraging diversity.

\paragraph{Self-Consistency Decoding} Given that LLM may generate non-unique or incorrect outputs for the same inputs, we use \emph{self-consistency} filtering \citep{self-consistency} to select pseudo-labels. Specifically, we create a consensus output for each unique input by selecting the most frequent answer; in the case of ties, we heuristically select the shortest answer.
This promotes accuracy of the generated dataset while ensuring unique examples.
\paragraph{Asynchronous Batching} API requests are parallelized using \emph{zeno-build} \citep{Neubig_Zeno_GPT_Machine_2023}. We use additional mechanisms, such as dynamic batch size and throttling, to optimize API usage.

\subsection{Model Retriever}
\label{sec:model_retriever_impl}
We need to select an appropriate model to finetune. To support many tasks with a unified model interface, we presently limit ourselves to encoder-decoder architectures on Hugging Face \citep{wolf-etal-2020-transformers}, following recent work that shows that encoder-decoder models are more data-efficient for model distillation \citep{calderon-etal-2023-systematic}.
This restriction still leaves a large set of pretrained models to choose from, e.g. \texttt{Salesforce/codet5-base} for coding-related tasks \citep{codet5} or \texttt{MaryaAI/opus-mt-ar-en-finetuned-ar-to-en} for Arabic-to-English translation \citep{opus_mt}. We frame the problem of selecting a pretrained model as a search problem. Using the user's instruction as a query, we search against all textual descriptions of models on Hugging Face.\looseness=-1

\begin{figure}[ht]
\begin{center}

    \includegraphics[width=0.4\textwidth]{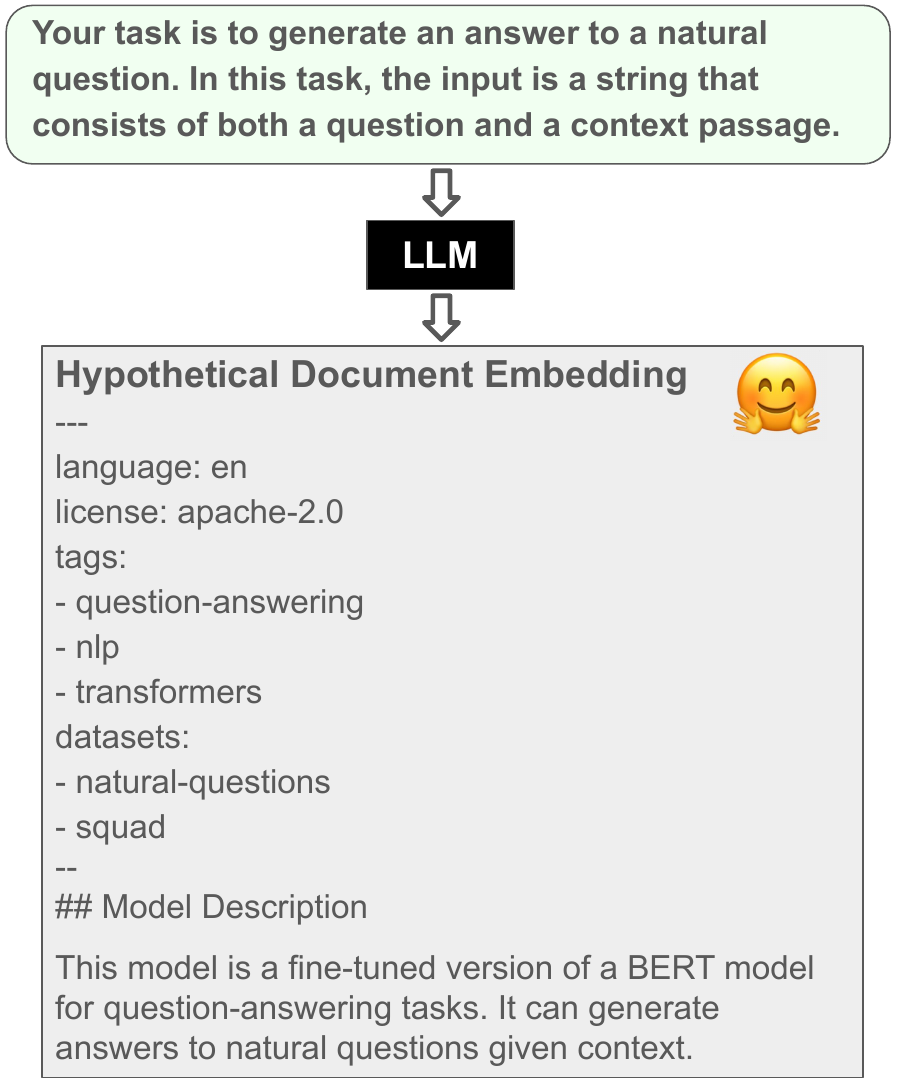}
    
\end{center}
\vspace{-10pt}
\caption{For our model retriever, we first construct a hypothetical model description for a query, then compute similarity scores between that hypothetical model description and the descriptions of real models.
}
\label{fig:hyde}
\vspace{-10pt}
\end{figure}

This search task is challenging because Hugging Face model descriptions are sparse and contain lots of templatic text, often with only a few words that signify the content of the model. To address this, we follow the HyDE framework \citep{gao-etal-2023-precise} and first use \texttt{gpt-3.5-turbo} to create a \emph{hypothetical model description} given the user's instructions. We show an example of a hypothetical document generated for a question-answering instruction in \autoref{fig:hyde}. 
Using this description as an expanded query, we then apply the BM25 algorithm to compute query-model similarity scores \citep{Robertson1995OkapiAT}. To ensure the ease of deployment of the resulting model, we filter out models whose size (in bytes) exceeds a user-specified threshold (set to 3GB by default). Using the intuition that highly-downloaded models are more likely to be high in quality, we choose the top model after ranking by:
\vspace{-5pt}
$$ BM25(\text{query}, \text{model})\cdot\log( \text{\# of Downloads} + 1).$$

\subsection{Training}
\label{sec:training_impl}

\paragraph{Dataset Processing}
We train the model by leveraging two datasets- one generated and one retrieved. To sidestep the challenge of making schema-specific modeling decisions (e.g. constructing specialized architectures for classification or generation tasks), we treat all datasets as ``text-to-text'' problems \citep{10.5555/3455716.3455856}. We textualize the input columns of each dataset and prepend the user's instructions to the input to guide the model.

\paragraph{Finetuning}
We concatenate the retrieved and generated datasets and shuffle them before training the student model. We use the same default hyperparameters for all tasks.\footnote{We empirically find that these default hyperparameters are effective, but we plan on implementing hyperparameter selection using generated validation data in the future.} We train with the AdamW optimizer with \texttt{lr = 5e-5} for 3 epochs, which takes roughly one hour for all tasks.

\subsection{Evaluation}
\label{sec:eval_impl}
Our \emph{Model Evaluator} automatically evaluates models for all tasks using three general-purpose metrics: \texttt{Exact Match}, \texttt{ChrF++} \citep{chrf}, and \texttt{BERTScore} \citep{BERTScore}. \texttt{ChrF++} balances precision and recall to assess text generation quality. \texttt{Exact Match} measures how often the model output perfectly matches the exact reference. \texttt{BERTScore} captures semantic similarities despite different wordings or phrasings by comparing the model output and reference in the embedding space. We use XLM-R \citep{conneau-etal-2020-unsupervised} as the encoder for BERTScore to support multilingual evaluation.

\subsection{Web App Creation}
\label{sec:web_app_creator_impl}
We finally provide an optional step in \promptmodel to automatically create a graphical user interface that allows downstream users to interact with the trained model. This web application, built using Gradio \citep{abid2019gradio}, can then be easily deployed publicly on a server.

\section{Experimental Setup}
\label{sec:experimental_setup}

\paragraph{Tasks}
As a proof-of-concept, we test our system's ability to learn a model for three tasks:

\begin{itemize}[nosep, leftmargin=1.1em,labelwidth=*,align=left]
\item \emph{Machine Reading Question Answering}: We first consider a common use case where pretrained models and training datasets are plentiful. We use SQuAD \citep{rajpurkar-etal-2016-squad} as ground truth to evaluate this setting.

\item  \emph{Japanese NL-to-Code}: Code generation from Japanese-language queries is a challenging scenario where prior work exists but no annotated data or pretrained models are available. We use MCoNaLa \citep{wang-etal-2023-mconala} for evaluation.

\item  \emph{Temporal Expression Normalization}: We finally consider a task where there are no pretrained models or training datasets of any kind available. Here we use the Temporal dataset of \citet{10.1145/3581641.3584059} as ground truth for evaluation.
\end{itemize}
Though \promptmodel offers automated model evaluation (on generated and retrieved datasests), we use real benchmark datasets here to measure our pipeline's ability to train accurate models.

\paragraph{LLM Baseline}
A primary goal of our work is to train small models that can match or outperform LLMs. To measure success towards this goal, we report the performance of \texttt{gpt-3.5-turbo} on each benchmark. We provide \texttt{gpt-3.5-turbo}\footnote{We used \texttt{gpt-3.5-turbo-0613}, accessed between July 26 and August 6, 2023.} the same instruction and demonstrations provided to \promptmodel, for fair comparison.

\section{Experiment Results}
\label{sec:results}
\subsection{Downstream performance}

How effective is \promptmodel at producing a high-quality model? In \autoref{tab:final_results}, we evaluated models produced by \promptmodel, as well as our baseline LLM \texttt{gpt-3.5-turbo}, on real benchmark datasets for each task --- SQuAD, MCoNaLa, and Temporal. 
We further examine the effect of removing 2 specific elements of the \promptmodel pipeline --- model retrieval and dataset retrieval.

\begin{table}[t]
\centering
\fontsize{9.3}{10}\selectfont

\begin{tabular}{r | c|c|c}
\toprule
 \textbf{Method} &  \textbf{SQuAD} &  \textbf{MCoNaLa} &  \textbf{Temporal}\\
 & {\color{gray}(EM)} & {\color{gray}(ChrF++)}  & {\color{gray}(ChrF++)} \\
\midrule
Prompt2Model & 61.5 & 13.1 & 55.2  \\
w/o Model Ret.  & 61.5 & 15.8 & 55.2  \\
w/o Data Ret. & 50.2 & 16.6 & N/A  \\
\midrule
\texttt{gpt-3.5-turbo} & 42.1 & 37.3 & 30.7 \\
\bottomrule
\end{tabular}
\caption{We evaluate the model produced by \promptmodel on real benchmarks for each test set, compared to \texttt{gpt-3.5-turbo}, which we used to power our dataset generator. We also examine the effect of removing specific parts of our pipeline --- model retrieval and dataset retrieval. There are no relevant datasets available for the Temporal task, so we did not use retrieved data for \promptmodel there.
}
\vspace{-10pt}
\label{tab:final_results}
\end{table}

On 2 of 3 datasets, we find that \promptmodel produces models that are considerably more accurate than \texttt{gpt-3.5-turbo}. This is remarkable because the retrieved model for SQuAD and Temporal is Flan-T5, which, at 250M parameters, is up to 700 times smaller than \texttt{gpt-3.5-turbo} (which is believed to contain 175B parameters).

We observe that \promptmodel's performance on MCoNaLa's Japanese-to-Python task is significantly worse than \texttt{gpt-3.5-turbo}. One explanation for this is the relatively low diversity in the generated dataset of Japanese queries; 45 of 5000 examples are different ways of saying ``find the maximum value in a list of numbers``. We do not observe this level of redundancy in our other datasets, suggesting that \texttt{gpt-3.5-turbo} may struggle to generate diverse text for non-English languages. Another reason is the lack of an appropriate student model --- the models found by the model retriever were trained on either on multiple language or code, but not both. The resulting pretrained models may lack the parametric knowledge to represent the Japanese inputs, Python outputs, or both.

\subsection{Combining retrieved and generated datasets is powerful}
Ideally, generated and retrieved data should be as close to the target domain as possible. In our experimental setting, where we deliberately choose prompts that mimic existing datasets, we can evaluate how well the model performs relative to a model trained on the same amount of data from the true dataset. We use SQuAD as a running example.\footnote{We focus on only SQuAD here because our other two tasks have less real training examples than the datasets we generate, making comparison impractical.} 
As our prompt is a description of the SQuAD passage-level question answering task (\autoref{fig:intro_diagram}), we exclude SQuAD from our retrieved datasets list. Instead, we evaluate models finetuned on:
\begin{enumerate}[nosep, leftmargin=1.2em,labelwidth=*,align=left]
    \item 3k examples from the closest retrieved dataset\footnote{The closest dataset retrieved by the dataset retriever for our SQuAD-inspired prompt is The Children's Book Test Dataset \cite{hill2016goldilocks}.}
    \item 3k examples generated by \promptmodel 
    \item The union of the above, which is what the full \promptmodel pipeline uses
    \item 3k examples from SQuAD (analogous to the user custom-annotating data for a task). %
\end{enumerate}

Table~\ref{tab:squad-comparison} shows the results across these four settings. While using retrieved or generated data causes a reduction in performance due to domain shift, the combination of the two methods achieves similar performance to using the true dataset. For this machine reading comprehension task where the user would need to custom-annotate data for their task, \promptmodel allows for \textit{similar performance at less than 1\% of the cost}. %

\begin{table}[]
\small
\centering
\setlength{\tabcolsep}{2.5pt}
\begin{tabular}{@{} r|c|c|c @{}}
\toprule
\textbf{Method} & \textbf{\#Train} & \textbf{Performance} & \textbf{Anno. Cost} \\ \midrule
Retrieval only                 & 3,000               &  56.79                        & $\approx$ \$ 0           \\
Generation only                & 3,000               & 44.20                    &  $\approx$ \$  5           \\
\textbf{Retrieval+generation}           & 6,000               & 61.46                    & $\approx$  \$  5           \\ 
\midrule
Custom annotation             & 3,000               & 61.64                    &  $\approx$ \$ 540     \\
\bottomrule
\end{tabular}
\caption{
We compare model performance on SQuAD on an annotation-cost basis, using datasets produced by different modules of \promptmodel, along with fully-manual annotation. 
Performance reported for all models is the exact match on the test set,\footnote{Exact match is the standard metric used for measuring SQuAD performance.} which reflects \emph{the true task performance}. 
Cost of custom annotation is estimated from \citet{rajpurkar-etal-2016-squad} using their reported annotator pay rate of \$9/hour and keeping 1,000 validation examples.
}

\label{tab:squad-comparison}
\end{table}

\subsection{Our generated evaluation data can identify real modeling improvements}

\hspace{-5pt}
\begin{table}[t]
\fontsize{9.5}{10}\selectfont
\centering
\begin{tabular}{r | r|c l}
\toprule
\textbf{Dataset} & \textbf{Metric} & \textbf{$\tau$} & \textbf{\(p\)-value}\\
\midrule
SQuAD & EM & 64.3 & 0.03* \\
Temporal & ChrF++ & 24.2 & 0.31  \\
MCoNaLa (JP) & ChrF++ & 70.9 & 0.00** \\
\bottomrule
\end{tabular}
\caption{We evaluate 10 different models on real test sets and their corresponding generated clones. We compute Kendall's Tau on the ranked lists of models and find statistically significant correlations for 2 of 3 datasets.}
\label{tab:kendall_results}
\end{table}

High-quality generated data should also allow us to \emph{discriminate} between multiple candidate models to select a model that will perform well downstream. We finetune various models on a generated dataset and rank their performance according to the generated test data and the test data from the target (real) dataset.  We evaluate the Kendall's rank correlation \cite{kendall1938new} between the two rankings to determine if our generated data can effectively determine which models are likely to perform well downstream. This is closely related to the concept of concurrence between benchmarks \cite{liu-etal-2023-question}; however, we are evaluating whether the generated and real data rank \textit{specific models} in the same ordering, rather than \textit{modeling approaches}. %

Table~\ref{tab:kendall_results} shows the Kendall's $\tau$ for each task, computed over a set of reasonable models.\footnote{This set of models consisted of 5 T5-family models, 2 BART-family models, and 1-5 additional retrieved models from the \textit{Model Retriever}, depending on task.} The generated data shows strong correlation to the true performance on two of the three datasets.

\section{Discussion and Conclusion}
\label{sec:conclusion}
We propose \promptmodel, a framework that automatically constructs task-specific models using only natural language prompts.
Our proof-of-concept experiments show that, despite using a similar easy-to-use interface as LLMs, \promptmodel delivers small yet accurate models and its generated datasets can be used to estimate real-world performance.
Besides our reference implementation providing a ready-to-use tool, \promptmodel's extensible design and modular implementation makes it a platform for advancing model distillation, dataset generation, synthetic evaluation, dataset retrieval, and model retrieval.

We believe our \promptmodel framework can inspire various novel research questions. We hope that our platform enables future work that looks more deeply into quality assurance on the generated data and the model. 
Interesting questions include how much data should we generate for downstream model training and how diverse should it be? How do we effectively mix the retrieved and generated dataset such to achieve complementary strengths (e.g. using dataset generation to focus on the expected inputs to the model that the retrieved dataset fails to cover)? 
Since users often struggle to articulate their needs up front, future extensions should also address the challenge of human-in-the-loop correction -- either by offering potential strategies to help humans iteratively refine prompts, or allowing humans to perform post-hoc fixes when the task metadata extraction and generated data do not align with their intentions. We hope to propose explicit challenges and invite the community to contribute novel implementations of various components in our framework.

\section*{Limitations}
One of the primary limitations of our system is that our current experiments have all been conducted using the \texttt{gpt-3.5-turbo} API (used for prompt parsing, dataset generation, and model retrieval). This LLM is paid and closed-source, which makes this problematic as a scientific artifact \citep{rogers2023closed}. Furthermore, the service provider of this LLM, OpenAI, prohibits the use of their API to create models that may compete with OpenAI, creating potential legal concerns with the use of \promptmodel in commercial applications. We are exploring the integration of open-source LLMs to avoid our reliance on proprietary APIs.

Another limitation of our work is the limited ability of \promptmodel to support tasks that require processing languages other than English. While we have shown the limitations of our system at supporting code generation from Japanese natural language queries, our system is likely to struggle more with lower-resource languages. We use the unpublished \texttt{gpt-3.5-turbo} model for our Dataset Generator in our reference implementation. This model is believed to be similar to GPT-3 \citep{NEURIPS2020_1457c0d6}, which was trained on 93\% English documents, 1\% German documents, 1\% French documents, and <5\% documents in any other language. Our use of this model may exacerbate existing disparities in language technologies between high-resource languages and low-resource languages.

One potential limitation is that we have only tested our approach on 3 tasks, each with a single dataset and a single evaluation metric. We justify this decision because our focus is on providing an extensible software system rather than establishing state-of-the-art results on many datasets, but we believe that our results suggest broader applicability.

\section*{Ethics Statement}

Any system which makes powerful technology more accessible to the public has ethical implications. \citet{deepfakes} discuss ethical issues with open-source packages in relation to software libraries for deepfaking, including the possibility of enabling malicious actors to use technology that they would otherwise not have the technical skills to leverage. This is also a risk for an AutoML system such as \promptmodel; however, we believe this risk is outweighed by the benefits of greater accessibility, especially given that a low barrier to entry for generating harmful data already exists in the form of prompted, web-interface models. 

While \promptmodel could, if given harmful inputs, generate toxic, offensive, or inaccurate synthetic data, this is no more of a risk with \promptmodel than it is with the underlying prompted model \cite{bender-parrots}; indeed, the use of models and supplementary datasets retrieved from Hugging Face may lessen the likelihood of a downstream model replicating harms from the prompted model's outputs, though more investigation is needed. Like all ML models, the models that \promptmodel returns can make mistakes, and we aim to be transparent in our documentation about potential limitations of the system. 

We hope that \promptmodel will be broadly useful. Our work is motivated by a desire to increase the accessibility of NLP models to people who are not in the NLP community but would benefit from the community's innovations; particularly, to people who would use NLP models downstream but may not have the domain-specific knowledge to design their own system. \promptmodel may also prove useful for early NLP researchers by providing a starting point for intuitions about baselines for various tasks and enabling the discovery of similarities between a described task and existing work. We open-source \promptmodel and welcome community contributions. 

\section*{Acknowledgements}
This work was supported in part by a fellowship from NEC Research Laboratories. We are grateful to Alex Cabrera,  Will Epperson, Nelson Liu, Arjun Ramani, Zirui Cheng, Zhiyuan Zeng, Tianci Xue, Yanchen Liu, Yi-Hsin Hung and Zhilin Yang for their feedback and guidance. We particularly appreciate Zirui Cheng's video production support for our demo.

\bibliography{anthology,custom}
\bibliographystyle{acl_natbib}

\end{document}